# Syllable based DNN-HMM Cantonese Speech-to-Text System

Timothy WONG, Claire W. Y. LI, Sam LAM, Billy CHIU, Qin LU, Minglei LI, Dan XIONG, Roy S. YU, Vincent T.Y. NG

Department of Computing, The Hong Kong Polytechnic University, Hong Kong
timtim.hk@connect.polyu.hk, {cswyli, cswolam, bchiu, csluqin, csmli,csdxiong, cssyu, cstyng}@comp.polyu.edu.hk

**Abstract**

This paper reports our work on building up a Cantonese Speech-to-Text (STT) system with a syllable based acoustic model. This is a part of an effort in building a STT system to aid dyslexic students who have cognitive deficiency in writing skills but have no problem expressing their ideas through speech. For Cantonese speech recognition, the basic unit of acoustic models can either be the conventional Initial-Final (IF) syllables, or the Onset-Nucleus-Coda (ONC) syllables where finals are further split into nucleus and coda to reflect the intra-syllable variations in Cantonese. By using the Kaldi toolkit, our system is trained using the stochastic gradient descent optimization model with the aid of GPUs for the hybrid Deep Neural Network and Hidden Markov Model (DNN-HMM) with and without I-vector based speaker adaptive training technique. The input features of the same Gaussian Mixture Model with speaker adaptive training (GMM-SAT) to DNN are used in all cases. Experiments show that the ONC-based syllable acoustic modeling with I-vector based DNN-HMM achieves the best performance with the word error rate (WER) of 9.66% and the real time factor (RTF) of 1.38812.

## 1. Introduction

In hybrid context dependent DNN-HMM speech recognition methods, an artificial neural network (ANN) with multiple non-linear hidden layers is trained to output posterior probabilities of output frame labels corresponding to tied HMM triphone states (senones). The input of a higher-dimensional feature vector is composed from consecutive concatenated frames in which Mel-Frequency Cepstral Coefficients (MFCC) or filter-bank features are successively normalized and transformed on a per speaker basis. While the alignments and output labels for DNN training come from the GMM-HMM system with a set of HMM triphone states and their corresponding Gaussian models. Due to power of DNN and its efficiency in learning discriminative features, recent attempts have used much higher dimensional input features to DNN-HMM systems such as the I-vector based speaker adaptive training (SAT) techniques, in which several hundred dimensional per-speaker I-vector features as extra input are added to the conventional 40 dimensional GMM-SAT features as the final input to the neural net (Karafiát et al., 2011; Gupta et al,, 2014). The GMM-SAT features are spliced across several to tens of consecutive frames rather than just one frame in a GMM system. Related works have been done using syllable-based acoustic modeling on large-vocabulary continuous speech recognition (LVCSR) for both monosyllabic and polysyllabic languages, including Mandarin (Lee et al., 1993; Pan et al., 2012; Deng Li and Li Xiao, 2013; Li et al., 2013; Hu et al., 2014; X. Li, and X. Wu, 2014) and West languages (Hinton et al., 2012; A. Mohamed, 2012; Swietojanski et al., 2013; Gupta & Boulianne, 2013; Schmidhuber 2015). However, automatic STT on Cantonese is far behind. This motivated us to investigate SST for Cantonese. In this work, all models are implemented using the Kaldi toolkit. The rest of the paper is organized as follows. Section 2 presents both the acoustic and language models used in our system. Section 3 describes the baseline DNN-HMM system which does not use I-vector for speaker adaptation, followed by the DNN-HMM system with speaker adaptation in Section 4. Experiments on the system-wide parameter tuning and performance evaluation will be discussed in Section 5. Section 6 reports the conclusion and the future works.

## 2. Acoustic Modeling, Language Modeling, and Lexicon

The key components in LVCSR are the acoustic model (AM) and the language model (LM). AM in a series of statistical DNN-HMM models parameterizes the statistical raw speech signal variations in phone-level sequences, while LM constraints the syntax and semantic meaningful word-level transcriptions accordingly using the context history. The lexicon used in our system contains 15,542 entries with 2,089 pronunciation variations as a bridge to map the word-level transcriptions to its phone (or clustered ones referred as state) sequence, and hence to a sequence of the context dependent states in AM given the context and HMM transducers.

### 2.1 Syllable-based Acoustic Modeling

Syllable-based modeling is pervasively used in West languages and Mandarin (Wu and Wu, 2007). Li (Li et al., 2013) achieved a significant improvement by replacing the traditional Initial-Final model with the IF triphone model in Mandarin.

#### 2.1.1. IF-based vs ONC-based Acoustic Units

In the conventional syllable-based modeling, "Initial+Final" units are regarded as the basic units of acoustic models. In Cantonese, there are 625 base syllables which can be divided into two phonological units: 19 initials (or onset) and 53 finals (LSHK, 1997) with 6 tones by default associated with finals.

Initials are typically consonants and each final consists of a vowel nucleus and a consonant coda. While the nucleus is indispensable, the onset and the coda are optional. The initials can be semi-vowel, nasals or non-nasal such as liquids, glides, fricatives, affricates, and plosives. For finals, they can be a vowel (long), a diphthong, a vowel with nasal coda, a vowel with stop coda, and syllabic nasal. Comparatively, Mandarin has only 23 initials and 37 finals. It is evident that recognizing Cantonese syllables faces more variation challenges than Mandarin due to the significant large number of syllables in Cantonese. Our study proposed that the conventional IF-based syllable units can be strengthened to model the intra-syllable variations of Cantonese by further breaking down the IF into "Initial/Onset + Nucleus and Coda" (ONC). Finals can actually be formed from a more elementary set of 15 nuclei plus 9 codas, a total of 24 elementary phonemes as depicted in Table 1. The 6 tones associated with the finals in IF will be allocated to both nucleus and codas to finely reflect the actual tone variations between nucleus and codas even within an individual character when combining with its context character(s) to form n-grams. For example, "令" should be encoded as "l i02 _ng01" in bi-gram "令狐" (one Chinese family name), yet as "l i02 _ng02" in "命令" (order). At the first comparison, the ONC scheme can halve the size of the phone set for modeling the finals, which is from 53 to 24. The ONC scheme is helpful in modeling modern Cantonese sound variations which the IF scheme is difficult to process due to its lack of flexibility. The IF scheme assumes that the 53 finals are distinctive, i.e. different finals give different Cantonese characters. However, due to language evolvement, some finals are becoming similar and hard to differentiate from each other. And even native Cantonese speakers find some of them hard to distinguish.

For example, the characters "baat" (八, eight) and "baak" (百, hundred) are no longer distinguished since both are pronounced as "baak", especially in youngsters. The two characters differ only in the ending unreleased stops, the alveolar "/t˥/" and the velar "/k˥/". Merging the coda "/k˥/" into "/t˥/" is also found following other nuclei, such as "got" (割, cut) and "gok" (各, each), or "bat" (筆, pen) and "bak" (北, north). The nasal finals are also being merged when speakers find the codas velar "/N/" the same as alveolar "/n/", such as in "san" (新, new) and "sang" (生, alive), "laan" (懶, lazy) and "laang" (冷, cold) respectively. These sound variations can undermine the basic assumption of 53 distinctive finals in the IF scheme because the finals are not distinctive in some characters. As this phenomenon is widespread in training corpus where expected individual corrections on the oral transcripts is infeasible, this results in unavoidable wrong mappings. For instance, when training the recognition of "baak" (百, hundred), instances of different pronunciations of "baat" (八, eight) are incorporated unavoidably. Similar cases also hold for the nasal codas. The influence of sound variations is minimized in the ONC scheme with more elementary building units of finals, i.e. nuclei and codas are extracted from the different finals and hence trained separately. In the IF scheme, for the same Final "aak" as in "baak", its training can only be confined to other instances with just different initial combinations, but still vulnerable to sound variations whereas the building units of "aak" in the ONC scheme are trained separately from different Onset-Coda and Onset-Nucleus combinations. For example, the vulnerable "/k˥/" can be trained from much larger correct instances in other speech environments, such as from "bik" (逼, coerce) which is never merged into "bit". This benefit cannot be obtained in the IF scheme, as it would be deemed to be "ik" and "aak" which are distinct training units. To summarize, the finer structure of the ONC scheme allows a deeper phoneme generalization from different speech environment and can alleviate the influence of sound variations. Inspired by Li's work on Mandarin (Li et al., 2013), we split the finals into nucleus and coda to model the intra-syllable variations, given that sub-syllables are deemed to have more stable acoustic realizations than IF (Wu and Wu, 2007).

| Base Syllable in IF/ONC | | | |
|---|---|---|---|
| **IF** | Initial | Final with tone | |
| 令狐 | l | ing4 | |
| 命令 | l | ing6 | |
| **ONC** | [Onset] | Nucleus with tone | [Coda with tone] |
| 令狐 | l | i02 | _ng05 |
| 命令 | l | i02 | _ng02 |

Table 1: Syllable for Cantonese-"令狐" and "命令".

### 2.1.2. Acoustic Data

Our acoustic model is trained by a subset of continuous sentence speech data from CUCorpora (Lee et al., 2002) which is a manually transcribed Cantonese speech with speaker identities so that speaker-adapted training can be conducted. The full utterance data is split into the training data, referred to as CUSent and the testing data, referred to as CUTest. Table 2 summarizes the training data CUSent which has a total of 21,576 utterances from 34 male and 34 female speakers, respectively. CUTest contains 1,198 utterances by 6 male and 6 female speakers.

| Speakers | 68 |
|---|---|
| Utterances | 21,576 |
| Syllables | 1,613 |
| Onsets | 20 |
| Nuclei | 15 |
| Codas | 9 |
| Initials | 20 |
| Finals | 53 |

Table 2: Summary of Speech Training Data (CUSent).

## 2.2 Language Modeling and Raw Corpus

Conventional count-based n-gram language modeling is used in lattice generation with first-pass decoding, and both n-gram modeling and Recurrent Neural Network language modeling (RNNLM) are used in lattice-based rescoring. For n-gram language modeling, we studied the Chinese Language Models with respect to either the character-level or the word-level (Mnih and Hinton, 2009; Mikolov et al., 2011b; Mikolov et al., 2011a) ones. Word-level language modeling is said to outperform character-level modeling (Luo et al., 2009) in Mandarin speech to text recognition. However, Chinese language in nature is character based in which the syllable rules constrain the syllable sequences and, if missed, cannot be revealed from word sequences merely. Furthermore, different segmentation algorithms and vocabularies included in the segmenting dictionary affect the recognition performance directly. Therefore, Hybrid character-word-level language modeling were used (Ng et al., 2008; Oparin et al., 2012; Liu et al., 2013) with significant improvement of up to 7.3% relatively (Liu et al., 2013). In general, language model interpolation techniques are classified into mixtures of experts (MoE) (Rosenfield, 1996) based linear model and products of experts ( PoE) (Rosenfeld et al., 2001) based log-linear model, and/or their combination. A multi-level language model interpolation technique is explored in Liu's study (Liu et al., 2013) using both combination models in different interpolation levels in which linear interpolation is used when interpolating over diverse text corpus and log-linear interpolation is used when interpolating character-level LMs into word-level LMs.

As our system is a domain oriented speech application system to aid local young students, our pre-trained experiments fixed a local online newspaper as our domain corpus. Developed through SRILM, a 10-year raw newspaper text in the size of 847M as domain corpus and another general corpus in the balanced size are processed in linear interpolation technique for a character-level LM generation to approximate the proposed syllable-based AM. The resulting 2-gram LM, referred to as LM2 of size 41.3M, is used in decoding for efficiency reasons. A large 4-gram LM, referred to as LM4 of size 603M, is used in lattice rescoring. Based on the work by (Gupta, V. and Boulianne, G. 2013) on WER reduction, rescoring using RNNLM is done to the top 200 entries. The RNNLM we used referred to as LM-RNNLM of size 39.6M is trained through Kaldi scripts using 100 hidden layers with 200 neurons in each layer. The vocabulary list obtained contains 37,950 higher frequent words.

## 3. DNN-HMM Systems

We set up two DNN acoustic models one without speaker adaption and another one with speaker I-vector. The two models are trained over either the IF-based modeling or the ONC-based models to evaluate the performances. Figure 1 shows the control flow of the diagram of the system. The two red boxes in Figure 1 correspond to speaker adaption using I-vectors and will be discussed in detail in Section 4.

### 3.1 Base-line System

In the baseline DNN-HMM system without I-vector features, the Acoustic Model (AM) is trained in two phases. Phase I involves building up a GMM-HMM system in which SAT technique based features are generated from raw wave frames with feature-space maximum likelihood linear regression (fMLLR), referred to as the GMM-SAT features. HMM model deals with the temporal variability of speech with each phone a strict left to right 3-state HMM with self-loop and the next state transition, while GMM estimates how well each state of each HMM fits a frame. As mentioned in Section 2, by using the lexicon, the word-level transcriptions are converted to its mapped phones (or states) sequences, and then to a sequence of the context dependent states with the corresponding context and HMM transducers. Each HMM state has its own GMM for this estimation. The context dependent triphone model in our system is cross-word based. Phase II involves a DNN training with the above features of several concatenating frames from Phase I as its input, which generates a label for each frame in the training set.

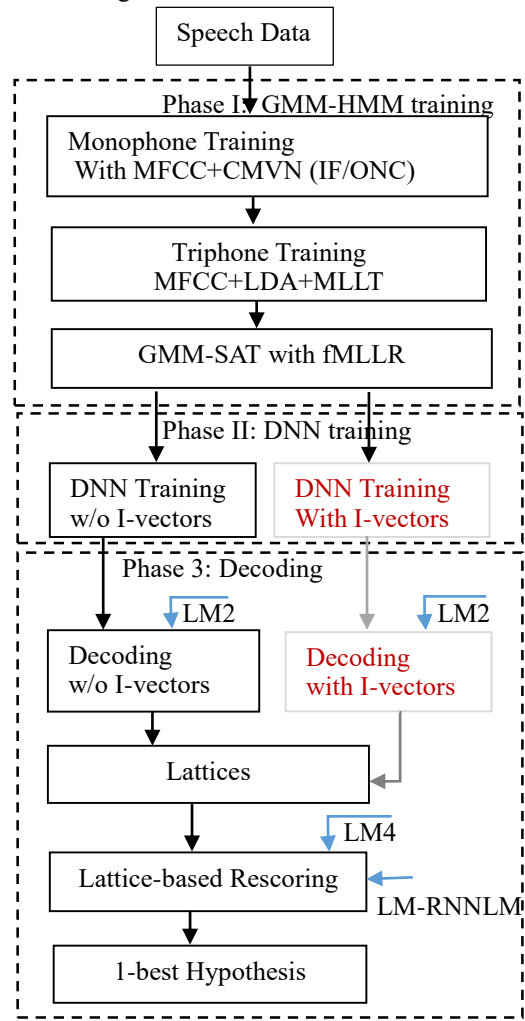

Figure 1: The Control flow of the System.

The input feature vector to DNN is generated as follows: 15 frames (7 left, 7 right) of the standard 13-dimentional MFCC features without energy are spliced together, then normalized by the cepstral mean-variance normalization (CMVN) to a zero mean vector and enhanced with its delta and delta-delta coefficients forming a 585-dimensional feature vector. Linear Discriminant Analysis (LDA) is applied to the resulting features for de-correlation and dimensionality reduction to project down to a vector of 40 dimensional features, and further de-correlation applied with Maximum Likelihood Linear Transform (MLLT) estimation. Then SAT is estimated on top of it with fMLLR adopted for speakers. It will be then forwarded to the input layer in Phase II which is a back-propagation DNN AM training.

In Phase II, we set up a 4 hidden layers p-norm nonlinearity neural network with a softmax output layer for the output posterior probabilities of 2,268 output frame labels corresponding to 2,268 context-dependent HMM states from Phase I. Figure 2 outlines the input and output DNN layers in Phase II. The red box is only needed with I-vector based speaker adaption. Note that each hidden layer has the size of 3,500 pnorm input dimensions and 350 pnorm output dimensions respectively. The initial learning rate starts from 0.01 for the first epoch, and the final learning rate is 0.001 for a total 15 training epochs. The mini-batches is set as 512 frames for weights updating during training. The 6-layer DNN has a total of 8.3 million weights. The training process was accelerated using one NVidia K2200 GPU on a single machine.

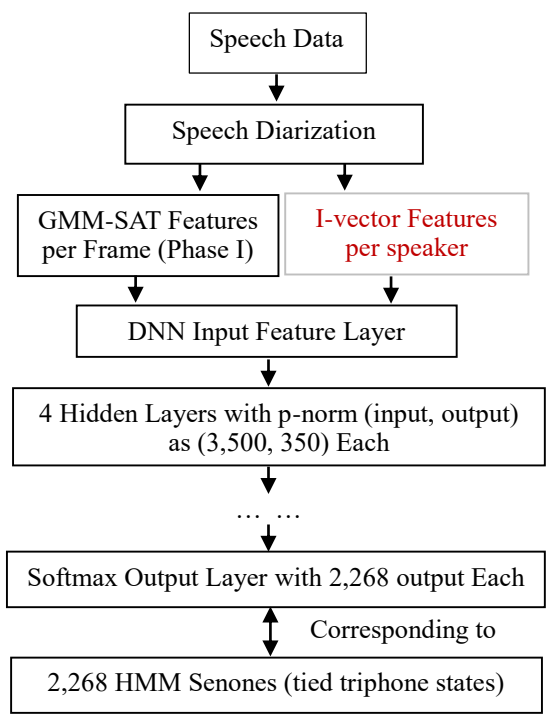

Figure 2: I-vector based DNN in Phase II.

Kaldi composes the context dependent HMM state-level AM with lexicon and LM to build a word-phone paired search graph in Weighted Finite-State Transducers (WFST) to recognize the test speech concurrently. In the DNN decoding phase (Phase III), the decoder will search the graph and LM2 will be used during decoding to generate the first best lattices with their alternative candidates. The lattice-based rescoring will be done twice using 4-gram LM4 and LM-RNNLM, respectively. Figure 3 shows a sample word-level lattice through search graph corresponding to the utterance of "合共九千九百萬元 (a total of nigh thousand and nigh hundred millions)". The output posterior probabilities over the state-level best path 0-1-13-14-5-6-7-11-12 is indicated in red.

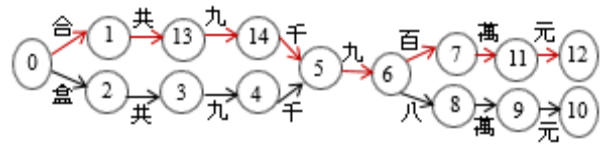

Figure 3: A sample lattice generated from the output layer through the decoding search graph.

### 3.2 DNN-HMM with I-Vector Adaptive Training

Speaker adaptive training technique tries to reduce the mismatch between the training and the testing data of different speakers, which can significantly improve the recognition performance with trained speaker information adapted. In the I-vector based DNN-HMM system, we trained I-vectors to identify speaker characteristics as additional features to DNN input. Both training and decoding use one I-vector per speaker. That is, all frames of an individual speaker have the same fixed dimensional (100 in our system) I-vector added to the input layer of the neural network. First, the Diagonal Universal Background Model Training (UBM) is applied on top of MFCCs and their first and second derivative features with an online CMVN applied, and then transformed with an LDA+MLLT matrix to train a diagonal mixture model of Gaussians with initial frames set to 400,000. The resulting diagonal Gaussian model is then used to train the I-vector extractor using on-line fashion from Kaldi for extracting I-vectors which has original raw features without Cepstral Mean Normalization (CMN) as input. The last step is to extract the optimized 100-dimensional I-vector from utterances for each speaker with the trained I-vector extractor.

I-vector-based DNN-HMM system takes the I-vector features plus GMM-SAT features as DNN input for I-vector based training and decoding as shown in Figure 1 in red boxes. Therefore, the input feature dimension is expanded by 100 tuned in the experiments for a trade-off RTF, particularly in our on-line STT system. The only difference in the two systems is an extra I-vector extractor training with more speaker characterized input features which is marked by the red box in Figure 2.

### 4. Experiments and Evaluation

The four configurations using a combination of either IF or ONC with or without I-vector based speaker adaption are evaluated by the standard character-based WER and decoding real time factor (RTF) which defines the ratio of recognition time in terms of the processing time to the duration of the speech input in terms of utterance length.

### 4.1 Performance Evaluation

Table 3 shows the evaluation results using CUSent as training data and CUTest as the testing data. The parameters used for Table 3 are system default values including: beam width = 15, max_actives_states = 7,000, I-vector dimension = 100, and the number of splicing frames for input to DNN = 15.

| Baseline | WER | RTF |
|---|---|---|
| IF (no I-vector) | 11.18% | 2.5361 |
| ONC (no-I-vector) | 10.39% | 1.30221 |
| IF (with I-vector) | 10.62% | 2.81336 |
| ONC (with I-vector) | 9.66% | 2.21266 |

Table 3: Performance evaluation on dataset (CUTest).

From Table 3 we can see that by adding speaker adaption using I-vector, both IF-based and ONC-based models will have improved WER with slight loss on RTF. To look at the data in more details, we can see that without the use of speaker adaption, the improvement of ONC-based compared to IF-based is 7.07%. This indicates that ONC-based is also more suited for Cantonese speech recognition. Improvement for using speaker adaptation for the ONC-based model gives another 7.03% improvement. The overall improvement by using ONC-based with I-vectors compared to IF-based without I-vectors can reach 13.60%, only slightly smaller than the improvement added together (7.07+7.03=14.10%). This indicates that the improvement brought out by the use of ONC-based model and I-vector based speaker adaption is almost orthogonal.

In terms of RTF, the ONC-based system is actually better than the IF-based one no matter speaker adaptation is used or not. It is not surprising, however, speaker adaptation will incur additional overhead if there is no tuning of system parameters.

### 4.2. Parameter Tuning

During decoding, two parameters, the max_actives_states and beam width are linked to both WER and RTF. The value of max_active_states defines the maximum number of active tokens in decoding, the smaller its value, the faster the recognizing (the smaller the RTF). Decoding beam width is used during graph search to prune hypotheses at state-level which when increasing decreases the WER and increases the RTF. From (Gupta, V. 2013), reducing the search beam width results in significant WER increase in GMM-HMM system. However, DNN-HMM is not very sensitive to beam width in French speech system. So, to further investigate the trade-off between WER and RTF by adjusting max-active-states and beam, we conducted two sets of experiments on max_active_states and beam by keeping all other parameters the fixed default values. Details of the experiment on beam adjustment are shown in Table 4 with max-active-states as 7,000. The experiments verified that the default beam value of 15 gives the best performance.

| Beam | WER | RTF |
|---|---|---|
| 13 | 9.78% | 1.89884 |
| 14 | 9.71% | 1.98288 |
| 15 | 9.66% | 2.21266 |
| 16 | 9.77% | 3.81283 |
| 17 | 9.78% | 4.8287 |

Table 4: Tuning of beam width.

| Max-active-states | WER | RTF |
|---|---|---|
| 2000 | 9.66% | 1.38812 |
| 3000 | 9.68% | 1.65792 |
| 5000 | 9.70% | 1.89007 |
| 7000 | 9.66% | 2.21266 |
| 9000 | 9.73% | 4.01179 |

Table 5: Tuning of max-active-states.

Table 5 shows the tuning of max-active-states given the frame width fixed at 15. Note that when max-active-states is reduced to 2000, the RTF is reduced to 1.38812 without sacrificing on WER which is at the lower rate of 9.66%, which means max-active-states only marginally affects WER but it is important to RTF. So the best parameters for WER at 9.66% and RTF at 1.38812 are tuned as: beam width = 15, max_actives_states = 2,000, I-vector dimension = 100, and the number of splicing frames for input to DNN = 15.

### 4.3. Error Analysis on IF and ONC models

Table 6 shows the number of correct and incorrect sentences for both IF-based and ONC-based systems with speaker adaptation. Out of the 1,198 number of sentences in CUTest, 650 sentences in IF-based modeling and 668 sentences in ONC-based modeling are clear of errors. Table 7 analysis the composition of the errors. Among the sentences bearing recognition errors, 489 of them are in both IF-based and ONC-based systems, 59 of them in IF-based one only, and 41 in ONC-based one only. Among the 489 sentences which have errors in both models, 286 of them have exactly the same errors as type 1 errors, labeled by [1] in Table 7. That is, 59.49% of all the shared errors are identical. 203 of them, referred to as Type 2 errors, labeled by [2] in Table 7, have different errors in the same sentence in 41.51% of the shared error set.

|  | IF | ONC |
|---|---|---|
| Correct Result | 650 | 668 |
| Incorrect Result | 548 | 530 |
| WER | 10.62% | 9.66% |

Table 6: Correct-Incorrect sentences in IF/ONC.

| Errors in IF only | 59 |
|---|---|
| Errors in ONC only | 41 |
| Errors in shared sentences (Identical) | (1)286 with exactly the same errors in 59.49% |
| | (2)203 with the different errors in 41.51% |

Table 7: The number of the identical and different error sentences in IF-ONC modeling.

Table 8 lists three sample sentences with type 1 errors in the first row and type 2 errors in the second and third rows over IF/ONC-based modeling.

| |
|---|
| Ref: CNF6F-285 該罐裝奶含天然乳糖 |
| (The canned milk contains natural lactose) |
| IF: CNF6F-285 該罐裝奶含天然魚塘 |
| ONC: CNF6F-285 該罐裝奶含天然魚塘 |
| Ref: CNF5F-439 日圓傾向上衝一二二水平 |
| (Japanese Ren is rising up to the level of 122) |
| IF: CNF5F-439 日圓偏向上衝一二二水平 |
| ONC: CNF5F-439 若沿天向上衝一二二水平 |
| Ref: CNFFM-007 但也不及日本人的愛月程度 |
| (Less favor of the moon as much as the Japanese) |
| IF: CNFFM-007 但也不及日本人的外語程度 |
| ONC: CNFFM-007 但也不及一般人的外語程度 |

Table 8: Sample sentences with type 1 and 2 errors.

Type 1 errors are mostly due the AM dealing with multiple identical or similar pronunciations so that recognizer has to rely on LM to choose the correct one. In the above sample, the frequency of '魚塘 (fishpond)' is 417 versus '乳糖 (lactose)' being 146 in the LM raw corpus. This example shows the limitation of conventional n-gram LM when handling long context history. Even if both of them are in the lexicon (as both are relatively frequent words), it will not help. If the long-distance dependency "奶 (milk)" can be successfully represented by the LM, the candidate hypothesis of "乳糖 (lactose)" would obtain a higher LM score than "魚塘 (fishpond)". In principle, RNNLM architectures can handle long distance relations and is better in dealing with data sparsity problem. RNNLMs have been applying in large scale acoustic modeling speech recognition systems for lattice rescoring (T. Mikolov et al., 2010; X. Liu et al., 2014; M Sundermeyer, 2015) given a lattice based search space in a DNNs hybrid framework, or for both the first pass decoding as well as the later lattice rescoring in the ASR system with RNNs architecture (H. Sak, A. Senior, and F. Beaufays, 2014; M Sundermeyer, 2015). For the ASR systems with RNNs, two main architectures of conventional recurrent and long sort-term memory (LSTM) neural networks are evaluated. LSTM RNNs are said to be more effective than DNNs and conventional RNNs, especially for speech recognition systems training and running on a single machine (H. Sak, A. Senior, and F. Beaufays, 2014). This value is also concerned in upcoming Kaldi's nnet3 framework with LSTM RNN architecture based on nnet2 framework.

For type 2 errors, different types of phonological modeling units provide different recognition results. This gives incentive to build systems which use multiple lattices based on multiple acoustic modeling to do a combined rescoring.

## 5. Conclusion and Future Works

The paper reports our work of an on-going automatic Cantonese speech recognition system with hybrid DNNs framework using Kaldi toolkit. We used the ONC-based syllable scheme which is derived from the IF-based scheme to improve the performance of AM from reducing the phonetic variations, combined with I-vector based speaker adaption, the overall performance can be improved by 13.60%. Compared to Mandarin, Cantonese is phonetically more challenging to process. Our work indicate that we can achieve WER at 9.66% with DNN architecture which is comparable to Mandarin at 7.93% in the recent report on noisy speech with deep RNN architecture (Amodei et al., 2015).

As future work, the LSTM RNN architecture as well as combined multi-lattice rescoring based on multiple acoustic modeling can be attempted to obtain further improvement.

## 6. Acknowledgements

The project is partially supported by the Hong Kong Government's ITF fund (ITS/072/14) and the Hong Kong Polytechnic University project RTVU. Our special thanks to all Kaldi community.

## 7. Main References

Dario Amodei, Rishita Anubhai, Eric Battenberg, Carl Case, Jared Casper, Bryan Catanzaro, Jingdong Chen, Mike Chrzanowski, Adam Coates, Greg Diamos, Erich Elsen, Jesse Engel, Linxi Fan, Christopher Fougner, Tony Han, Awni Hannun, Billy Jun, Patrick LeGresley, Libby Lin, Sharan Narang, Andrew Ng, Sherjil Ozair, Ryan Prenger, Jonathan Raiman, Sanjeev Satheesh, David Seetapun, Shubho Sengupta, Yi Wang, Zhiqian Wang, Chong Wang, Bo Xiao, Dani Yogatama, Jun Zhan, Zhenyao Zhu. (2015). Deep Speech 2: End-to-End Speech Recognition in English and Mandarin (Baidu). *arXiv:1412.5567.*

Wenping Hu, Yao Qian, Frank K. Soong. (2014). A DNN-based Acoustic Modeling of Tonal Language and Its Application to Mandarin Pronunciation Training. *In Proceeding of IEEE International Conference on Acoustic, Speech and Signal Processing (ICASSP)*, pp. 3230-3234.

V. Gupta, and G. Boulianne. (2013). Comparing Computation in Gaussian Mixture and Neural Network Based Large Vocabulary Speech Recognition. In *14th*

*Annual Conference of the International Speech Communication Association*. Lyon, France.

V. Gupta, P. Kenny, P. Ouellet, and T. Stafylakis. (2014). I-Vector Based Speaker Adaptation of Deep Neural Networks for French Broadcast Audio Transcription. *Proc. ICASSP, Florence, Italy.*

Hinton, G., Deng, L., Yu, D., Dahl, G., Mohamed, A., Jaitly, N., Senior, A., Vanhoucke, V., Nguyen, P., Sainath, T., et al. (2012). Deep neural networks for acoustic modeling in speech recognition. *IEEE Signal Processing Magazine.*

Martin Karafiát, Lukáš Burget, Pavel Matějka, Ondřej Glembek and Jan Černocký (2011). iVector-Based Discriminative Adaptation for Automatic Speech Recognition. *In Proceedings of IEEE ASRU*, pp. 152-157. ISBN 978-1-4673-0366-8.

Lin-shan Lee, Chiu-Yu Tseng, Keh-Jiann Chen, I-Jung Hung, Ming-Yu Lee, Lee-Feng Chien, Yumin Lee, Renyuan Lyu, Hsin-Min Wang, Y-C Wu, et al. (1993). Golden mandarin (II) - an improved single-chip real-time mandarin dictation machine for Chinese language with very large vocabulary. *In Proc. ICASSP-93.* volume 2, pp. 503–506.

Tan Lee, Wai Kit Lo, PC Ching, and Helen Meng. (2002). Spoken language resources for Cantonese speech processing. *Speech Communication*, 36(3):327–342.

Deng. Li and Li Xiao. (2013). Machine Learning Paradigms for Speech Recognition: An Overview. IEEE Transactions on Audio, Speech, and Language Processing. 21(5), pp. 1060-1089.

X. Li, and X. Wu, (2014). *Labeling unsegmented sequence data with DNN-HMM and its application for speech recognition*. In ISCSLP, pp.10-14.

Xiangang Li, Caifu Hong, Yuning Yang, and Xihong Wu. (2013). Deep neural networks for syllable based acoustic modeling in Chinese speech recognition. *Signal and Information Processing Association Annual Summit and Conference (APSIPA),* pp. 1–4. IEEE.

X. Liu, Y. Wang, X. Chen, M. J. F. Gales, and P. C. Woodland. (2014). Efficient lattice rescoring using recurrent neural network language Models. *In Proc. ICASSP.* pp. 4941–4945.

Jun Luo, Lori Lamel, and J Gauvain. 2009. Modeling characters versuswords for Mandarin speech recognition. In *Acoustics, Speech and Signal Processing, IEEE International Conference on*, pp. 4325–4328. IEEE.

LSHK. (1997). Hong kong jyut ping characters table.

T. Mikolov, *M Karafiát*, L. Burget, J Cernocký and S. Khudanpur. (2010). Recurrent neural network based language model. *In Proceedings of INTERSPEECH, vol. 2010, no. 9. International Speech Communication Association*. pp. 1045–1048.

Tomas Mikolov, Anoop Deoras, Stefan Kombrink, Lukas Burget, and Jan Cernockỳ. (2011a). Empirical evaluation and combination of advanced language modeling techniques. *In INTERSPEECH*, number 1, pp. 605–608.

T. Mikolov, A. Deoras, D. Povey, L. Burget, J. Cernocky. (2011b). Strategies for Training Large Scale Neural Network Language Models. *Proc. ASRU-2011*, pp. 196–201.

Andriy Mnih and Geoffrey E Hinton. (2009). A scalable hierarchical distributed language model. *In Advances in neural information processing systems*, pp. 1081–1088.

A. Mohamed, George E Dahl, and Geoffrey Hinton. (2012). Acoustic modeling using deep belief networks. *Audio, Speech, and Language Processing, IEEE Transactions on*, 20(1):14–22.

H. Sak, A. Senior, and F. Beaufays. (2014). Long short-term memory recurrent neural network architectures for large scale acoustic modeling. *In Fifteenth Annual Conference of the International Speech Communication Association (INTERSPEECH). ISCA.*

M Sundermeyer, H Ney, R Schlüter. (2015). From feedforward to recurrent LSTM neural networks for language modeling. *IEEE/ACM Transactions on Audio, Speech and Language Processing (TASLP).* 23(3), pp. 517-529.